\documentclass{SIP}

\usepackage{amsmath}
\usepackage{amsthm}
\usepackage{hyperref}
\usepackage{graphics}
\usepackage{algorithmic}
\usepackage{subfig}
\usepackage{url}

\usepackage{epstopdf, epsfig}

\usepackage{multirow}
\usepackage{makecell}
\usepackage{tabulary}
\usepackage{booktabs}

\begin{document}

\title[Vision and Language]{Vision and Language: from Visual Perception to Content Creation}

\author[Tao Mei, Wei Zhang, Ting Yao]{Tao Mei, Wei Zhang, Ting Yao}

\address{\add{1}{JD AI Research, Building A, North-Star Century Center, 8 Beichen West Road, Beijing, China}}

\corres{\name{Tao Mei}
\email{tmei@jd.com}} 

\begin{abstract}
  Vision and language are two fundamental capabilities of human intelligence. Humans routinely perform tasks through the interactions between vision and language, supporting the uniquely human capacity to talk about what they see or hallucinate a picture on a natural-language description. The valid question of how language interacts with vision motivates us researchers to expand the horizons of computer vision area. In particular, ``vision to language" is probably one of the most popular topics in the past five years, with a significant growth in both volume of publications and extensive applications, e.g., captioning, visual question answering, visual dialog, language navigation, etc. Such tasks boost visual perception with more comprehensive understanding and diverse linguistic representations. Going beyond the progresses made in ``vision to language," language can also contribute to vision understanding and offer new possibilities of visual content creation, i.e., ``language to vision." The process performs as a prism through which to create visual content conditioning on the language inputs. This paper reviews the recent advances along these two dimensions: ``vision to language" and ``language to vision." More concretely, the former mainly focuses on the development of image/video captioning, as well as typical encoder-decoder structures and benchmarks, while the latter summarizes the technologies of visual content creation. The real-world deployment or services of vision and language are elaborated as well.
\end{abstract}


\maketitle

\section{Introduction}
Computer Vision (CV) and Natural Language Processing (NLP) are two most fundamental disciplines under a broad area of Artificial Intelligence (AI). CV is regarded as a field of research that explores the techniques to teach computers to see and understand the digital content such as images and videos. NLP is a branch of linguistics that enables computers to process, interpret and even generate human language. With the rise and development of deep learning over the past decade, there has been a steady momentum of innovation and breakthroughs that convincingly push the limits and improve the state-of-the-art of both vision and language modelling. An interesting observation is that the research in the two area starts to interact and many previous experiences have shown that by doing so can naturally build up the circle of human intelligence.

\begin{figure*}[!tb]
\vspace{-0.0in}
	\centering {\includegraphics[width=0.99\textwidth]{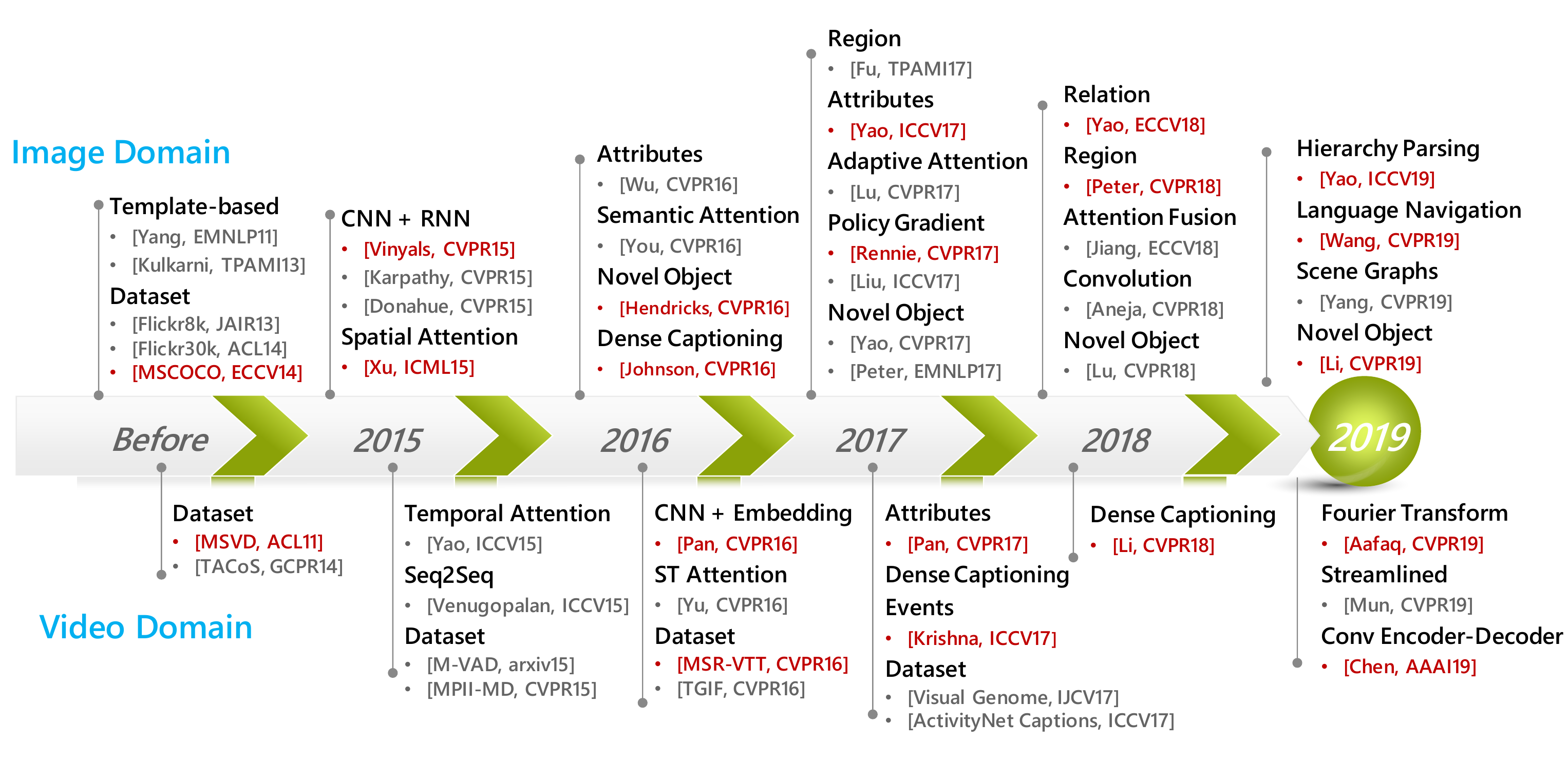}}
\vspace{-0.0in}
	\caption{A road map for the techniques and datasets in vision (image/video) to language in 10 years.}
	\label{fig:VL1}
\end{figure*}

In general, the interactions between vision and language have proceeded along two dimensions: vision to language and language to vision. The former predominantly recognizes or describes the visual content with a set of individual words or a natural sentence in the form of tags \cite{yao2013annotation}, answers \cite{anderson2018bottom}, captions \cite{yao2017boosting,yao2018exploring,yao2019hierarchy} and comments \cite{li2016share}. For example, a tag usually denotes a specific object, action or event in visual content. An answer is a response to a question about the details depicted in an image or a video. A caption goes beyond tags or answers by producing a natural-language utterance (usually a sentence) and a comment is also a sentence which expresses an emotional state on visual content. The latter of language to vision basically generates visual content according to natural language inputs. One typical application is to create an image or a video from text. For instance, given a textual description of ``this small bird has short beak and dark stripe down the top, the wings are a mix of brown, white and black," the goal of text-to-image synthesis is to generate a bird image which meets all the details.

This paper reviews the recent state-of-the-art advances of AI technologies which boost both vision to language, particularly image/video captioning, and language to vision. The real-world deployments in the two fields are also presented as the good examples of how AI transforms the customer experiences and enhances user engagement in industrial applications. The remaining sections are organized as follows. Section \ref{sec:v2l} describes the development of vision to language by outlining a brief road map of key technologies on image/video captioning, distilling a typical encoder-decoder structure, and summarizing the evaluations on a popular benchmark. The practical applications of vision to language are further presented. Section \ref{sec:l2v} details the technical advancements on language to vision in terms of different conditions and strategies for generation, followed by a summary of progresses on language to image, language to video, and AI-empowered applications. Finally, we conclude the paper in Section \ref{sec:conclusion}.

\section{Vision to Language}\label{sec:v2l}
This section summarizes the development of vision to language (particularly image/video captioning) in several aspects, ranging from the road map of key techniques and benchmarks, typical encoder-decoder architectures, to the evaluation results of representative methods.

\subsection{Road Map of Vision to Language}
In the past 10 years, we have witnessed researchers strived to push the limits of vision to language systems (e.g., image/video captioning). Fig. \ref{fig:VL1} depicts the road map for the techniques behind vision (image/video) to language and the corresponding benchmarks. Specifically, the year of 2015 is actually a watershed in captioning. Before that, the main stream of captioning is template-based method \cite{kulkarni2013babytalk,yang2011corpus} in image domain. The basic idea is to detect the objects or actions in an image and integrate these words into pre-defined sentence templates as subjective, verb and objective. At that time, most of the image captioning datasets are ready to use, such as Flickr30K and MSCOCO. At the year 2015, deep learning-based image captioning models are first presented. The common design \cite{Vinyals14} is to employ a Convolutional Neural Network (CNN) as an image encoder to produce image representations and exploit a decoder of Long Short-Term Memory (LSTM) to generate the sentence. The attention mechanism  \cite{Xu:ICML15} is also proposed at that year which locates the most relevant spatial regions when predicting each word. After that, the area of image captioning is growing very fast. Researchers came up with a series of innovations, such as augmenting image features with semantic attributes \cite{yao2017boosting} or visual relations \cite{yao2018exploring}, predicting novel objects through leveraging unpaired training data \cite{yao2017novel,li2019novel}, and even going a step further to perform language navigation \cite{wang2019reinforced}. Another extension direction of captioning in image domain is to produce multiple sentences or phrases for an image, aiming to recapitulate more details within image. In between, dense image captioning \cite{johnson2016densecap} and image paragraph generation \cite{wang2019convolutional} are typical ones, which generate a set of descriptions or paragraph that describes image in a finer fashion.

\begin{figure*}[!tb]
\vspace{-0.0in}
 \centering {\includegraphics[width=0.97\textwidth]{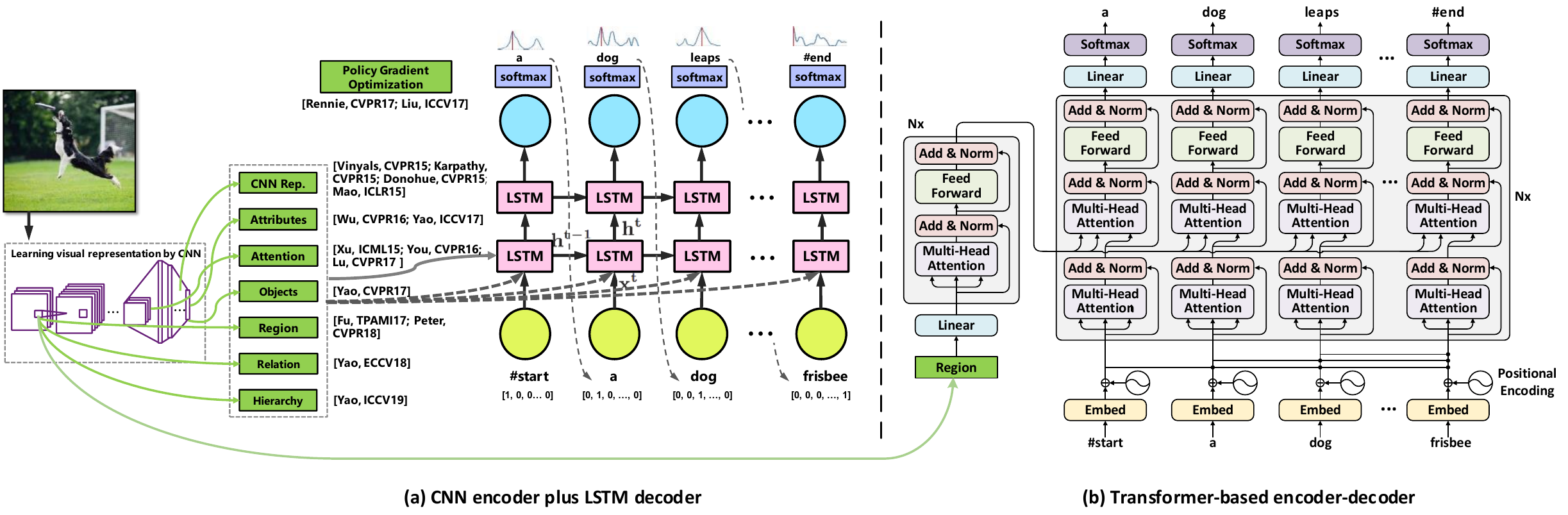}}
\vspace{-0.0in}
 \caption{The typical architectures of (a) CNN encoder plus LSTM decoder and (b) Transformer-based encoder-decoder for image captioning.}
 \label{fig:VL2}
\end{figure*}

\begin{table*}[!tb]
\centering
\caption{The reported performance (\%) of image captioning on COCO testing server with 5 reference captions (c5) and 40 reference captions (c40).}
\begin{tabular}{lccccccccc}
\Xhline{0.8pt}
\multirow{2}*{Model} & \multirow{2}*{Group} & \multicolumn{2}{c}{B@4} & \multicolumn{2}{c}{METEOR} & \multicolumn{2}{c}{ROUGE-L} & \multicolumn{2}{c}{CIDEr-D}\\
\cmidrule(lr){3-10}
{} &{} &c5&c40&c5&c40&c5&c40&c5&c40\\
\hline
HIP \cite{yao2019hierarchy}  &JD AI, ICCV'19 &\textbf{39.3}	&\textbf{71}	&\textbf{28.8}	&\textbf{38.1}	&\textbf{59}	&\textbf{74.1}	&\textbf{127.9}	&\textbf{130.2}\\
GCN-LSTM \cite{yao2018exploring} &JD AI, ECCV'18 &38.7	&69.7	&28.5	&37.6	&58.5	&73.4	&125.3	&126.5\\
RFNet \cite{jiang2018recurrent} &Tencent, ECCV'18 &38	&69.2	&28.2	&37.2	&58.2	&73.1	&122.9	&125.1\\
Up-Down \cite{anderson2018bottom}	&MSR, CVPR'18	&36.9	&68.5	&27.6	&36.7	&57.1	&72.4	&117.9	&120.5\\
LSTM-A \cite{yao2017boosting}	&MSRA, ICCV'17	&35.6	&65.2	&27	&35.4	&56.4	&70.5	&116 &118\\
Watson Multimodal \cite{rennie2017self}	&IBM, CVPR'17 	&35.2	&64.5	&27.0	&35.5	&56.3	&70.7	&114.7	&116.7\\
G-RMI \cite{Liu:2016PGSPIDEr}	&Google, ICCV'17	&33.1	&62.4	&25.5	&33.9	&55.1	&69.4	&104.2	&107.1\\
MetaMind/VT-GT \cite{Xiong2016MetaMind}	&Salesforce, CVPR'17	&33.6	&63.7	&26.4	&35.9	&55	&70.5	&104.2	&105.9\\
reviewnet \cite{Yang:NIPS2016reviewnet}	&CMU, NIPS'16	&31.3	&59.7	&25.6	&34.7	&53.3	&68.6	&96.5	&96.9\\
ATT \cite{you2016image}	&Rochester, CVPR'16	&31.6	&59.9	&25	&33.5	&53.5	&68.2	&94.3	&95.8\\
Google \cite{Vinyals14}	&Google, CVPR'15	&30.9	&58.7	&25.4	&34.6	&53	&68.2	&94.3	&94.6\\
\Xhline{0.8pt}
\end{tabular}
\label{Table:VL}
\end{table*}

The start point of captioning in video domain is also in the year of 2015. Then, researchers start to remould the CNN plus RNN captioning framework towards the scenario of captioning in video domain. A series of techniques (e.g., temporal attention, embedding, or attributes) are explored to further improve video captioning. Concretely, \cite{yao2015describing} is one of early attempts that incorporate temporal attention mechanism into captioning framework by learning to attend to the most relevant frames at each decoding time step. \cite{pan2016jointly} integrates LSTM with semantic embedding to preserve the semantic relevance between video content and the entire sentence. \cite{pan2017video} further augments captioning model to emphasize the detected visual attributes in the generated sentence. It is also worthy mentioned that in 2016, MSR-VTT video captioning dataset \cite{Xu:CVPR16} is released which has been widely used and already downloaded by more than 100 groups worldwide. Most recently, \cite{aafaq2019spatio} applies short Fourier transform across all the frame-level features along the temporal dimension to fuse all frame-level features into video-level representation and further enhances video captioning. Another recent attempt for video captioning is to speed up the training procedure by fully employing convolutions in both encoder and decoder networks \cite{chen2019temporal}. Nevertheless, considering that videos in real life are usually long and contain multiple events, the conventional video captioning methods generating only one caption for a video in general will fail to recapitulate all the events in the video. Hence the task of dense video captioning \cite{krishna2017dense,li2018dense} is introduced recently and the ultimate goal is to generate a sentence for each event occurring in the video.

\begin{figure*}[!t]
\begin{center}
\epsfxsize =16cm
\epsffile{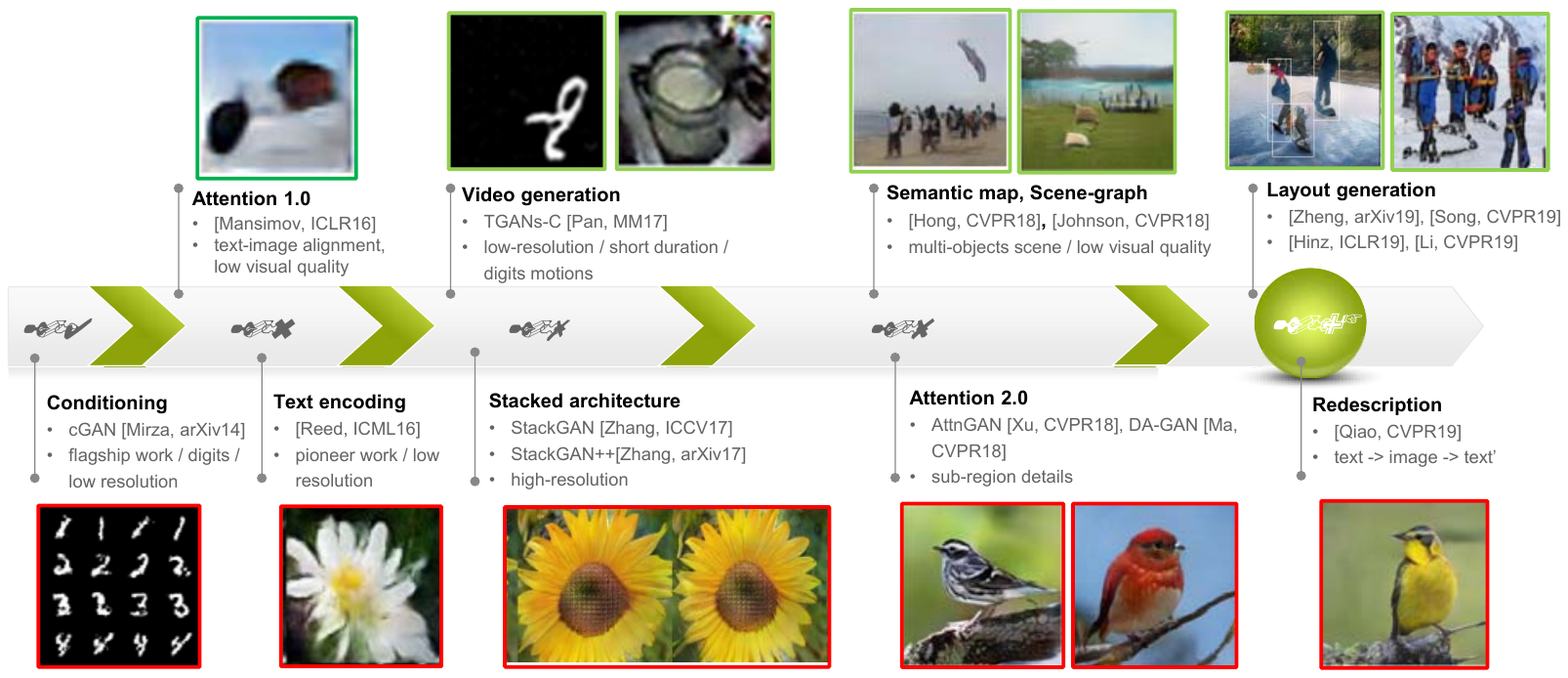}
\caption{The road map of ``language-to-vision" in past five years, while milestone techniques are marked along the year axis. Top: single object generation. Bottom: multiple-objects scene generation.}
\label{fig:roadmap}
\end{center}
\vspace{-0.15in}
\end{figure*}

\subsection{Typical Architectures}
According to the road map of vision to language, the mainstream of modern image captioning follows the structure of CNN encoder plus LSTM decoder, as shown in Fig. \ref{fig:VL2} (a).
In particular, given an image, image features can be firstly extracted through multiple ways: 1) directly taking the outputs of fully-connected layers as image features \cite{Vinyals14}; 2) incorporating high-level semantic attributes into image features \cite{yao2017boosting}; 3) performing attention mechanism to measure the contribution of each image region \cite{Xu:ICML15}; 4) extracting region-level features \cite{anderson2018bottom} and further exploring relation \cite{yao2018exploring} or image hierarchy \cite{yao2019hierarchy} on the region-level features. The image features will be further fed into LSTM decoder to generate the output sentence, one word at each time step. In the training stage, the next word is generated based on the previous ground-truth words while during testing the model uses the previously generated words to predict the next word. In order to bridge the mismatch between~training and testing, reinforcement learning \cite{rennie2017self,Liu:2016PGSPIDEr,ren2017deep} is usually exploited to directly optimize LSTM decoder with the sentence-level reward, such as CIDEr or METEOR.

Taking the inspiration from the recent successes of Transformer self-attention networks \cite{vaswani2017attention} in machine translation, recent attention has been geared toward exploring Transformer-based structure \cite{sharma2018conceptual} in image captioning. Fig. \ref{fig:VL2} (b) depicts the typical architecture of Transformer-based encoder-decoder. Different from CNN encoder plus LSTM decoder that capitalizes on LSTM to model word dependency, Transformer-based encoder-decoder model fully utilizes attention mechanism to capture the global dependencies among inputs. For encoder, $N$ multi-head self-attention layers are stacked to model the self-attention among input image regions. The decoder contains a stack of $N$ multi-head attention layers, each of which consists of a self-attention sub-layer and a cross-attention sub-layer. More specifically, the self-attention sub-layer is firstly adopted to capture word dependency and the cross-attention sub-layer is further utilized to exploit the co-attention across vision (image regions from encoder) and language (input words).

Similar to the mainstream in image captioning, the typical paradigm in video captioning is also essentially an encoder-decoder structure. A video is first encoded into a set of frame/clip/shot features via 2D CNN \cite{Vinyals14} or 3D CNN \cite{qiu2017learning,tran2015learning}. Next, all the frame-level, clip-level or shot-level visual features are fused into video-level representations through pooling \cite{pan2016jointly}, attention \cite{yao2015describing} or LSTM-based encoder \cite{Venugopalan:ICCV15}. The video-level features are then fed into LSTM decoder to produce a natural sentence.

\subsection{Evaluation and Applications}

\textbf{Evaluation}. Here we summarize the reported performance of representative image captioning methods on the testing server of popular benchmark COCO \cite{coco} in Table \ref{Table:VL}. In terms of all the evaluation metrics, GCN-LSTM \cite{yao2018exploring} and HIP \cite{yao2019hierarchy} lead to performance boost against other captioning systems, which verifies the advantage of exploring relations and hierarchal structure among image regions.

\textbf{Applications}.
Recently, there exist several emerging applications which involve the technology of vision to language. For example, captioning is integrated into online chatbot \cite{tran2016rich,pan2017seeing} and an ai-created poetry \cite{zhou2018design} is published in China. In JD.com, we utilize captioning techniques for personalized product description generation last year, which aims to produce compelling recommendation reasons for billions of products automatically.

\section{Language to Vision}\label{sec:l2v}

This section discusses from another direction of ``language to vision", \textit{i.e.}, visual content generation guided by language inputs. In this section, we start by reviewing the road map development, as well as the technical advancements in this area. Then we discuss the open issues and applications particularly from the perspective of industry.

\textbf{Visual Content Generation}. We briefly introduce the domain of visual generation, since ``language to vision" is deeply rooted in the same techniques. Over the past few years, we have witnessed great progresses in visual content generation.
The origin of visual generation dates back to  \cite{goodfellow2014generative}, where multiple networks are jointly trained in an adversarial manner. Subsequent works generate images in specific domains such as face \cite{pggan,Chen_2019_ICCV,styleGAN}, person \cite{ma2017pose,fashionGen,ma2018disentangled}, as well as generic domains \cite{brock2018large,pmlr-v97-lucic19a}. From the perspective of inputs, the generation can also be treated as conditioning on different information, e.g., noise vector \cite{goodfellow2014generative}, semantic label \cite{mirza2014conditional}, textual captions \cite{reed2016generative}, scene-graph \cite{johnson2018image} and images \cite{pix2pix2016,CycleGAN2017}. Among all these works, visual generation based on natural languages plays one of the most promising branches, since semantics are directly incorporated into the pixel-wise generation process.

\subsection{Road Map of Language to Vision}

Fig.~\ref{fig:roadmap} summarizes recent development of ``language to vision". In general, both the vision and language modalities are becoming more and more complicated, and the results are much more visually convincing, compared to when it was firstly introduced in 2014.

The fundamental architecture is based on a conditional generative adversarial network, where the conditioning input is usually the encoded natural language. After a series of transposed-convolutions, the language input is gradually mapped to a visual image with higher and higher resolution. The key challenges are in two folds: 1) how to interpret the language input, i.e., language representation, and 2) how to align the visual and textual modalities, i.e., the semantic consistency between vision and language. Recent results on single object (bottom) have already been visually plausible to human perception. However, state-of-the-art models are still struggling in generating scenes with multiple objects interacting with each other.

\subsection{Technical Advancements}

The success in language to vision generation are mostly based on the following technical advancements, which have become standard practices commonly accepted by the research community.

\textbf{Conditioning Input}.
Following the standard GAN framework~\cite{goodfellow2014generative}, \citet{mirza2014conditional} derived the conditional version GAN, which allows visual generation according to language inputs. The conditioning information can be in any forms of language, such as tag, sentence, paragraph, image, scene-graph and layout. Almost all subsequent works in ``language to vision" are based on the conditioning architecture. However at that time, only MNIST \cite{mnist98} digits are demonstrated in low resolution, and the conditioning information is merely a digit-label.

\textbf{Text Encoding}.
GAN-INT-CLS~\cite{reed2016generative} is the first work based on natural-language inputs. For the first time, it bridges the gap from natural language sentences to image pixels. The key step is based on learning a text representation based on a recurrent network to capture visual clues. The rest is mostly following \cite{mirza2014conditional}. Additionally, a matching-aware discriminator is proposed to keep the consistency between the generated image and textual input. Though the results still look primitive, people can draw flower images by altering textual inputs.

\textbf{Stacked Architecture}.
Another big advancement is by stackGAN \cite{zhang2016stackgan,stackgan++}, where stacked generators are introduced for high-resolution image generation. Different from previous works, stackGAN can generate realistic 256x256-pixel images by decomposing the generator into multiple stages stacked sequentially. The Stage-I network only sketches the primitive shape and color of the object based on text representation, yielding a low resolution image. The Stage-II network further fills details, such as textures, conditioning on the Stage-I result. A conditioning augmentation technique is also introduced to augment the textual input and stabilize the training process. Compared to \cite{reed2016generative}, the visual quality is much improved based on this stacked architecture. Similar idea is also adopted in Progressively-Growing GAN \cite{pggan}.

\textbf{Attention Mechanism}.
As in other vision tasks, attention is effective in highlighting key information. In ``language to vision", attention is particularly useful in aligning keywords (language) and image patches (vision) during the generation process. Two generations (v1.0 and v2.0) of attention basically follow this paradigm, but differs in many details, e.g., network architecture, text encoding. Attention 1.0, AlignDraw \cite{alignDraw}, proposes to iteratively paint on a canvas by looking at different words at different stages. However the results were not promising at that time. Attention 2.0, AttnGAN \cite{xu2018attngan} and DA-GAN \cite{dagan}, basically follows the similar paradigm, but improves significantly on image quality, e.g., fine-grained details.

\textbf{Semantic Layout}.
Recent studies \cite{fashionGen,Dong2018SoftGatedWF,bau2018gandissect} have demonstrated the importance of semantic layout in image generation, where layout acts as the blue-print to guide the generation process. In language to vision, semantic layout and scene-graph are introduced to reshape the language input with more semantics. \citet{hong2018inferring} propose to generate object bounding-boxes first, and then refine by estimating appearances inside each box. \citet{johnson2018image} encode objects relationship from scene graph to construct the layout for decoder generation with graph convolutions. \citet{zheng19} introduce spatial constraint module and contextual fusion module to model the relative scale and offset among objects for commonsense layout generation, and \citet{hinz19} further propose an object pathway for multi-objects generation with complex spatial layouts.

\subsection{Progress and Applications}

The development of ``Language to Vision" can be summarized as follows. On one hand, the language description is becoming more complex, i.e., from simple words to long sentences. On the other hand, the vision part is also becoming more complex, where objects-interaction and fine-grained detail are expected:
\begin{itemize}
    \item Language: label $\rightarrow$ sentence $\rightarrow$ paragraph $\rightarrow$ scene graph
    \item Vision: single object $\rightarrow$ multiple objects
\end{itemize}

\textbf{Language to Image}.
Early studies mainly focus on simple words and single-object images, e.g., birds \cite{cub200}, flowers \cite{flowers} and generic objects \cite{ILSVRC15}. As shown in Fig.~\ref{fig:roadmap} (bottom), the visual quality is much improved over the past few years, and some results are plausible enough to deceive human~eyes.

Though single-object image can be well generated, multi-objects scene still struggles for realistic results, as in Fig.~\ref{fig:roadmap} (top). A general  trend is to reduce the complexity by introducing semantic layout as an intermediate representation. Roughly, machines now can generate spatially reasonable images, but fine-grained details are still far from satisfactory at current stage.

\textbf{Language to Video}.
Compared to image, language to video is more challenging due to huge volume of information and extra temporal constraint. There is only a few works studying this area. For example, \citet{mmPan} attemp to generate video out of captions based on 3D convolution operation. However, the results are quite limited for practical applications.

\textbf{Applications}.
The application of ``language to vision" can be roughly grouped into two categories: generation for human eyes or for machines. In certain domains (e.g., face), language to vision already starts to produce highly plausible results with industrial standards\footnote{https://thispersondoesnotexist.com/}. For example, people can generate royalty-free facial photos on demand\footnote{https://github.com/SummitKwan/transparent\_latent\_gan} for games \cite{shi2019facetoparameter} or commercials, by manually specifying gender, hair, eyes. Another direction is generating data for machine and algorithms. For example, NVIDIA \cite{zheng2019joint} proposed a large-scale synthetic dataset (DG-Market) for training person Re-ID models. Also some image recognition and segmentation models start to benefit from machine-generated training images.
However, it is worth noting that despite the promising results, there is still a large gap for massive deployment in industrial products.

\section{Conclusion}\label{sec:conclusion}
Vision and language are two fundamental systems of human representation. Integrating the two in one intelligent system has long been an ambition in AI field. As we have discussed in the paper, on one hand, vision to language is capable of understanding visual content and automatically producing a natural-language description, and on the other hand, language to vision is able to characterize the intrinsic structure in vision data and create visual content according to the language inputs. Such interactions, while still at the early stage, motivate us to understand the mechanisms in connecting vision and language, reshape real-world applications, and re-think the end result of the~integration.


\bibliographystyle{plainnat}
\bibliography{reference}

\vskip2pc

\end{document}